\documentclass[10pt]{article}
\usepackage[preprint]{tmlr}

\usepackage{booktabs}
\usepackage{array}
\usepackage[font=small,labelfont=bf,labelsep=period]{caption}
\usepackage{float}
\usepackage{graphicx}
\usepackage{amsmath}
\usepackage{amssymb}
\usepackage{hyperref}
\usepackage{url}
\usepackage{microtype}
\newcommand{\artifactnote}[1]{
  \par\smallskip
  \noindent{\footnotesize\textit{Note.}\ #1\par}
  \medskip
}
\hypersetup{pdftitle={Retrieved-Span Training for Efficient Query-Focused Meeting Summarization on QMSum},pdfauthor={Edward Xi Yang},hidelinks}
\renewcommand{\headrulewidth}{0pt}

\title{Retrieved-Span Training for Efficient Query-Focused Meeting Summarization on QMSum\thanks{Agentic coding tools supported implementation, experiment execution, and prose drafting under the author's direction. The author remains responsible for all content; a fuller disclosure appears before the references.}}
\author{\name Edward Xi Yang \email edward@ertas.ai \\
      \addr Ertas AI \\
      Melbourne, Australia}

\begin{document}
\maketitle

\begin{abstract}
Query-focused meeting summarization requires answering from transcripts that exceed small-model context windows. QMSum provides no scorer, making reported results difficult to compare. We rescore or generate 15 systems under one implementation and release per-query predictions. Our central result concerns alignment between training and inference inputs. Through a common inference port, a released 406M Fusion-in-Decoder specialist loses 6.30 ROUGE-1 when moved from capped long input to 2,000-word retrieved spans. Fine-tuning it on this span regime recovers the loss. On test it scores 36.33 ROUGE-1 versus 35.41 for our 1.2B system; the meeting-cluster 95\% interval for the difference is [-0.27, +2.22], so QMSum does not statistically separate them. The 406M system uses about one-third as many total parameters and less than half the peak inference memory. Within the fixed 1.2B base, span-regime fine-tuning adds 5.29 [+4.02, +6.56], while replacing the first 4,500 transcript words with 2,000 retrieved words adds 1.55 on test and 0.29 on validation. Separately, under one concise prompt and reference-overlap scorer, a released 406M specialist exceeds five proprietary hosted models by at least 6.2 ROUGE-1; verbose outputs and absent human or factuality evaluation limit this ordering. More retrieval recall does not yield a measurable ROUGE gain at fixed budget, and gold spans add 1.09 [-0.22, +2.43]. We report ROUGE-1, ROUGE-2, ROUGE-L, ROUGE-Lsum and BERTScore with paired intervals. Conclusions are limited to QMSum and automatic metrics. We release the protocol, scorer, predictions and trained artifacts.
\end{abstract}

\section{Introduction}

Query-focused meeting summarization asks a model to answer a specific question from a transcript that is far longer than a small model can usefully attend to. QMSum transcripts average about 9,000 words and reach 25,000, while reference summaries have a median length of 59 words. One option is to send the whole transcript to a long-context hosted model. Another is to locate the relevant text and summarize only that fraction on local hardware.

Direct comparison is difficult because \textbf{QMSum ships no evaluation code}. Its repository contains data, span annotations, one system's outputs and a preprocessing notebook, but no scorer or stated ROUGE implementation; the paper says only ``standard ROUGE F-1''. Follow-up work from the same group uses the Perl implementation, whereas recent work generally uses Python reimplementations. The resulting scores are separated by an unknown implementation offset.

We therefore present a positioning result, not a leaderboard claim. We generate and score five zero-shot proprietary hosted models, four community checkpoints, a dialogue-pretrained specialist, our 1.2B locate-then-summarize system and its ablations, and a 406M segment-encoding specialist before and after retrieval-regime training. All rows share one scorer and at most one test-split touch. The promoted system has one labelled locator-side deviation from the otherwise frozen protocol (Section 3.2).

\textbf{The first ordering is metric- and prompt-specific.} Under one concise prompt and our reference-overlap scorer, a 406M Fusion-in-Decoder specialist from 2023 exceeds every proprietary hosted model we ran, by 6.2 ROUGE-1 over the strongest. Because its authors released both predictions and their score, we can also show that our implementation reproduces their result to within 0.12 ROUGE-1. Hosted answers are longer than the references, and we do not infer a human or factuality ordering from this comparison.

\textbf{The second ordering is about matching training to inference.} Through our inference port, the released specialist loses 6.3 ROUGE-1 when moved from its capped long-input regime to our retrieved spans. Span-regime fine-tuning on data byte-identical to our summarizer's removes that deficit. On test, the 406M model scores 36.33 ROUGE-1 and our 1.2B system scores 35.41. Their difference is not statistically separated under any reported metric; the meeting-cluster ROUGE-1 interval is [-0.27, +2.22]. The smaller model uses about one-third as many total parameters, less than half the peak inference memory, and about 45 minutes of training. The experiment does not establish whether span training beats the authors' own full-input pipeline, because our port has a measured 3.3-point offset (Section 8.3).

Two controls locate the gain. Within the fixed 1.2B base model, replacing the first 4,500 transcript words with 2,000 retrieved words adds 1.55 ROUGE-1 on test and 0.29 on validation, whereas span-regime fine-tuning adds 5.29 [+4.02, +6.56]. Replacing retrieved spans with gold spans adds 1.09 [-0.22, +2.43] on validation, near the benchmark's detection floor, and still leaves a 1.40-point residual to the specialist [+0.05, +2.76]. Training supplies most of the measured gain; retrieval remains useful chiefly as an input and cost constraint.

\textbf{Contributions.}

\begin{enumerate}

\item \textbf{A common scale} with one scorer, at most one test touch per system, explicit protocol deviations, and released per-query predictions.

\item \textbf{A metric-scoped specialist ordering:} under one prompt and scorer, a 2023 406M model exceeds every proprietary hosted model tested, with cross-implementation scorer calibration and explicit output-length caveats.

\item \textbf{A matched-regime comparison:} a span-trained 406M encoder-decoder is not statistically separated from a 1.2B decoder-only model on the test split while using fewer measured resources; fixed-model controls identify fine-tuning as the larger lever.

\item \textbf{Reusable measurement results:} encoder-context-aware retrieval windows, evidence that recall and ROUGE are decoupled at the tested operating point, and a roughly one-point full-split detection floor.

\end{enumerate}

\section{Related work}

\subsection{Query-focused meeting summarization}

Query-focused summarization conditions a summary on an information need rather than producing a generic digest. QMSum \citep{zhong2021qmsum} instantiates the task over 232 transcripts from academic (ICSI), product-design (AMI) and parliamentary meetings. The paper counts 1,808 query-summary pairs; the release contains 1,810 (Section 3.1). Transcripts average roughly 9,000 words and reach 25,000, and specific queries include human-annotated relevant spans, supporting both retrieval and an oracle.

Zhong et al. also introduce locate-then-summarize: select relevant transcript text, then summarize it. They pair a span locator with HMNet \citep{zhu2020hmnet} and report a gold-span oracle.

\subsection{The published QMSum landscape}

The strongest published QMSum results do not decompose the input. The 406M Segment Encoder encodes overlapping chunks independently and lets a BART-large decoder cross-attend over their concatenated representations \citep{vig2022segenc}, reaching 37.05 ROUGE-1 or 37.80 with Wikisum pre-finetuning. Question-driven pretraining raises the same architecture to 38.06 \citep{pagnoni2023socratic}; QontSum's contrastive objective reaches 38.42 \citep{sotudeh2023qontsum}, and LTRSum's segment-level ranking decoder reaches 38.82 \citep{sotudeh2024ltrsum}. Closest to our pipeline, \citet{liu2023ranker} rank utterances before generation and reach 35.51 with the same cross-encoder family we use.

More recently, CachED trains chunked encoder-decoder models end to end using gradient caching \citep{saxena2025cached}. It reports QMSum test ROUGE-1/2/L of 38.9/14.0/24.6 with 406M BART-large and 38.4/13.5/24.4 with 139M BART-base. These results reinforce the importance of matching long-input training to inference, but they are quoted rather than merged into Table 1: the paper uses the Perl ROUGE package, its own generation protocol, and a different BERTScore backbone, and we do not have its per-query predictions for rescoring.

Our system does not beat this line: it sits \textbf{3.2 ROUGE-1} below Pagnoni et al. under our scorer. On the 237 specific validation queries, a perfect locator closes \textbf{44\%} of the corresponding gap as a point estimate, with the lever's interval crossing zero (Section 7.3). The published system encodes about 15,700 tokens through a 406M encoder-decoder trained on long input; ours retrieves 2,000 words for a QLoRA-adapted 1.2B decoder-only model.

Through our inference port, moving their released checkpoint from capped long input to our spans costs 6.30 ROUGE-1 within model; at 29.00, that off-regime checkpoint is also 6.39 below our 35.39 validation row. Retraining it for that regime eliminates the measured deficit. On test, the two systems are not statistically separated under any reported metric, while the smaller system uses about one-third as many total parameters and less than half the peak inference memory (Section 7.5). The controlled intervention identifies input-regime training, rather than architecture alone, as the main source of the recovery.

Neither system is guaranteed a whole meeting. Their cap is about 11,800 words, binding on the longest transcripts (median test length 9,206; maximum 25,244), while our 2,000-word cap always binds. The contrast is therefore between two capped regimes with roughly 6x different budgets.

\subsection{Extract-then-generate and joint training}

SUMM\textasciicircum{}N summarizes coarse-to-fine \citep{zhang2022summn}. DYLE jointly trains extraction and generation, treating snippets as latent and aligning extraction with decoder attention \citep{mao2022dyle}. DialogLM and DialogLED instead pretrain long-dialogue models with window denoising and hybrid sparse attention \citep{zhong2022dialoglm}.

DYLE motivates a limitation: our locator and summarizer are trained separately, and we do not measure the value of joint optimization (Section 8.7).

\subsection{Retrieval versus long context}

At larger scales, a 4,000-token model with retrieval can match a 16,000-token extension at less compute \citep{xu2024retrieval}, whereas sufficiently resourced long context can lead on quality while retrieval retains a cost advantage \citep{li2024ragorlc}. We ask the adjacent small-model question on inputs well beyond the model's usable context.

\subsection{Parameter-efficient adaptation and small specialised models}

Quantized LoRA \citep{hu2022lora,dettmers2023qlora} makes our 1.2B model trainable at 6,144 tokens within 16 GB. Small fine-tuned models already outperform much larger zero-shot models in text classification \citep{bucher2024finetuned,gondara2025smallorlarge}. Section 6 gives a generation-side instance: a rescored 406M specialist exceeds every proprietary hosted model we ran under the shared prompt and scorer, while other published specialists point the same way but are not scorer-commensurable (Section 6.3).

\subsection{Summarization evaluation}

ROUGE \citep{lin2004rouge} and BERTScore \citep{zhang2020bertscore} are reference-anchored proxies. Token-overlap metrics correlate weakly and inconsistently with human judgement, especially among top systems \citep{fabbri2021summeval,bhandari2020reevaluating,kryscinski2019critical}. On extractive meeting summaries, \citet{liu2008rouge} likewise find weak ROUGE-human correlation, improved by handling disfluencies and speaker information. On QMSum specifically, \citet{kirstein2024metrics} correlate nine automatic metrics with a meeting-error taxonomy and find mostly weak to moderate relationships, including cases where metrics mask observable errors. Section 8.1 therefore scopes our claims and points to concurrent proposition-level evaluation.

Two points bear directly on our protocol. ROUGE-Lsum is distinct from ROUGE-L, so we report both (Section 3.3). Small test sets also make state-of-the-art comparisons underpowered \citep{card2020power}; on QMSum, 50-example slices give a 5.6-point ROUGE-1 interval (Section 3.4).

\section{Task, data, and protocol}

\subsection{Data}

We use the QMSum release without modification: 1,257 training, 272 validation and 281 test query-summary pairs, 1,810 total. The paper reports 279 test instances and 1,808 total, but the files we score, and the released predictions in Section 5.4, contain 281.

The test split covers 35 meetings, with 37 general and 244 specific queries. We normalize the specific-query annotations into utterance-index ranges; this normalized field is our construction.

\subsection{Protocol, and why it is frozen}

A committed module fixes seed 20260723, at most 512 new tokens, greedy decoding, a 900-word retrieval window, a 3,000-word span budget, a 40,000-word hosted-model transcript budget, and the Appendix A prompt. The transcript budget exceeds both split maxima (25,244 test; 24,573 validation), so no hosted-model input is truncated.

Validation drives all selection; test is used once per reported system. Section 7 therefore uses validation except for the span-trained SegEnc's declared test touch in Section 7.5.

\textbf{The promoted configuration has one labelled locator-side protocol deviation:} a 375-word window and 2,000-word span budget replace 900 and 3,000. It also uses a 12-layer rather than 6-layer ranker, which was not frozen; Section 4.1 lists all differences. We retain the committed module and label the deviation. Scorer, splits, seed, generation cap, decoding and test discipline are unchanged, and no baseline uses these locator values. Table 1 reports both the protocol-exact 33.39 and promoted 35.41 ROUGE-1 rows; released predictions record their settings.

\subsection{Scoring}

Because QMSum provides no scorer, we generate and score every comparison ourselves. We report ROUGE-1, ROUGE-2, ROUGE-L and ROUGE-Lsum F-measure with Porter stemming enabled, using \texttt{rouge-score==0.1.2}, and BERTScore F1 using \texttt{bert-score==0.3.13} with \texttt{roberta-large}. ROUGE-L and ROUGE-Lsum appear as separately labelled columns and never merged into one called ROUGE-L. Lsum requires sentence boundaries, which we derive with a single deterministic rule applied identically to candidate and reference, after flattening any line breaks the system emitted. Without that normalization, Lsum degenerates to L for text without newlines and rewards formatting. Rescoring moved Lsum by up to 4.2 points for line-breaking systems and left ROUGE-1, ROUGE-2 and ROUGE-L unchanged.

\textbf{Every between-system margin carries a paired bootstrap interval}, except the labelled plateau mean and worst-checkpoint curve summaries in Section 7.5. For per-query scores $a_1, \dots, a_n$ and $b_1, \dots, b_n$ from two systems on the same $n$ queries, we draw $R$ resamples of the query indices with replacement and take the difference within each draw:

\begin{equation*}
\delta^{(r)} \;=\; \frac{1}{n} \sum_{t=1}^{n}
  \Big( b_{\,j^{(r)}_t} \;-\; a_{\,j^{(r)}_t} \Big),
\qquad
j^{(r)}_1, \dots, j^{(r)}_n \ \sim\ \mathrm{Unif}\{1, \dots, n\},
\qquad r = 1, \dots, R
\end{equation*}

We report the 2.5th and 97.5th percentiles of $\{\delta^{(r)}\}$, with $R = 10{,}000$. Jointly resampling queries cancels shared between-query difficulty; an unpaired interval would be wider.

We quote published numbers from other papers only with their protocol stated, never in a column beside our own rows. Section 6 reports one measured exception to the usual pessimism about cross-implementation comparability.

\subsection{Evaluation variance on this benchmark}

\label{sec:variance}

Five measurements bound what this benchmark can resolve.

\textbf{Slice size.} Bootstrapping 10,000 seeded 50-example draws from our 281 test predictions gives a 95\% interval of \textbf{[32.75, 38.30] around a full-split ROUGE-1 of 35.41}, a span of \textbf{5.6 ROUGE-1 points}, roughly $\pm$2.8 either side. The same computation on our earlier configuration gave [30.60, 36.34] around 33.39, a span of 5.7, so the width is a property of the benchmark rather than one system. It exceeds most adjacent Table 1 gaps and our deficit to the specialist, so 50-example samples can reorder the table by chance. We therefore score complete splits.

\textbf{Full-split detection floor.} A paired bootstrap over all 281 test queries still gives a 95\% interval of roughly $\pm$0.9 to $\pm$1.4 ROUGE-1. Differences below about one ROUGE-1 point are undetectable on this benchmark at full split size, and the published QMSum results in Section 2.2 are separated by margins in exactly that band.

\textbf{Meeting-level clustering.} Because 281 queries nest within 35 meetings, query-level resampling assumes imperfect independence. A 10,000-draw meeting-cluster bootstrap widens the intervals without changing conclusions: +2.98 against gpt-5.6-luna becomes [+1.39, +4.63], the 3.19-point deficit to the specialist [+2.01, +4.47], and the specialist's 6.2 over the strongest hosted model [+4.84, +7.45], all still excluding zero, while the +0.93 span-trained SegEnc comparison becomes [-0.27, +2.22], still crossing it. The query-level intervals reported elsewhere are tighter and rest on this independence assumption.

\textbf{Checkpoint choice.} Five checkpoints from one run (epochs 4 to 8) span \textbf{1.7 ROUGE-1}; the peak is 1.4 above the next three's mean. This within-run spread is comparable to the detection floor, so a single checkpoint is not a reliable system estimate (Section 7.5).

\textbf{Training seed.} Changing only the seed moves validation ROUGE-1 by \textbf{1.59 points} (35.39 to 33.80, two runs), enough to erase the configuration gain in Section 6.2. Only one of four further attempted seeds completed (Section 8.9), so this is a two-run bound, not a distribution.

\textbf{Wall clock.} Identical reruns produced byte-identical predictions but moved mean latency by 17.2\% and 8.5\%. We therefore report named-run ranges, not fine-grained speed claims.

These standards change our reporting in both directions. We withdrew an earlier +0.96 ROUGE-1 hosted-model claim because its interval crossed zero [-0.30, +2.24]; the promoted +2.98 margin excludes it [+1.75, +4.24]. For the competitor in Section 7.5, we report the last checkpoint rather than the peak, reducing its winning margin by 1.3 ROUGE-1.

\section{System}

The system locates a transcript subset, then generates from that subset.

\subsection{Stage 1: the locator}

We chunk each transcript into fixed-width windows over speaker-labelled utterances, score every window against the query with a cross-encoder, and pack the highest-scoring windows in transcript order up to a word budget. Writing that out, for a transcript chunked into windows $W_1, \dots, W_m$ of width $w$ words and a query $q$:

\begin{equation*}
\begin{aligned}
s_i &= f_\theta(q, W_i),
& s_{i_1} &\ge s_{i_2} \ge \dots \ge s_{i_m}, \\
k &= \max\Big\{\, t \in \{1, \dots, m\} :
  \sum_{r=1}^{t} |W_{i_r}| \le B \,\Big\}, \\
\mathcal{S} &= \{\, i_1, \dots, i_k \,\}.
\end{aligned}
\end{equation*}

where $(i_1,\ldots,i_m)$ orders windows by score, $|W|$ is word length and $B$ is the span budget. Selected windows are restored to transcript order before generation. Equal widths, except for a final remainder, make greedy packing score-optimal up to that boundary case.

Section 3.2 explains why the best row is labelled as a deviation:

\begin{table}[H]
\centering
\small
\setlength{\tabcolsep}{5.5pt}
\begin{tabular}{>{\raggedright\arraybackslash}p{0.320\linewidth}>{\raggedright\arraybackslash}p{0.253\linewidth}>{\raggedright\arraybackslash}p{0.253\linewidth}}
\toprule
 & protocol-exact & promoted \\
\midrule
window width & 900 words & \textbf{375 words} \\
span budget & 3,000 words & \textbf{2,000 words} \\
ranking cross-encoder & \texttt{ms-marco-MiniLM-L-6-v2}, 6 layers, 22.7M & \textbf{\texttt{ms-marco-MiniLM-L-12-v2}, 12 layers, 33M} \\
training pairs & 13,738 (16.6\% positive) & \textbf{33,655 (11.9\% positive)} \\
utterance recall at its own budget (validation) & 0.627 & 0.670 \\
test ROUGE-1 of the full system & 33.39 & \textbf{35.41} \\
\bottomrule
\end{tabular}
\end{table}

Both rankers are fully fine-tuned by regressing query-window overlap with annotated spans. Window width is load-bearing because the 512-token encoder sees only a truncated view of a median 1,150-token, 900-word window (Section 7.1).

Against dense retrieval over the same windows, protocol-exact cross-encoder window recall at 3,000 words is 0.612 on validation and 0.628 on test, versus 0.549 for dense retrieval. Across validation budgets of 2,000 / 3,000 / 4,000 words, their curves are 0.490 / 0.612 / 0.704 and 0.433 / 0.549 / 0.665.

\textbf{When window width varies, we report utterance recall} because window recall's denominator changes with width. For annotated utterances $G$:

\begin{equation*}
R_{\mathrm{utt}} \;=\;
  \frac{\big|\, G \cap \bigcup_{i \in \mathcal{S}} W_i \,\big|}{|G|},
\qquad
R_{\mathrm{win}} \;=\;
  \frac{\big|\{\, i \in \mathcal{S} \ :\ W_i \cap G \neq \emptyset \,\}\big|}
       {\big|\{\, i \ :\ W_i \cap G \neq \emptyset \,\}\big|}
\end{equation*}

Narrower windows split the same gold content and enlarge the denominator of $R_{\mathrm{win}}$, whereas $|G|$ is fixed. At one configuration the measures are close (0.6122 window versus 0.6268 utterance recall for the protocol-exact locator on validation).

The locator takes 22 ms per protocol-exact query and 71 ms per promoted query, 0.8\% and 2.9\% of test latency. Its 0.1 to 0.3 GB activation peak is small beside the system's 5.7 GB validation peak. A perfect locator adds \textbf{1.09 ROUGE-1}, near the detection floor (Section 7.3).

\subsection{Stage 2: the summarizer}

The summarizer is \texttt{LiquidAI/\allowbreak{}LFM2.\allowbreak{}5-\allowbreak{}1.\allowbreak{}2B-\allowbreak{}Instruct}, adapted on 1,095 specific-query targets with 4-bit QLoRA (rank 16, alpha 32, dropout 0.05, all linear layers), learning rate 2e-4, three epochs, 6,144-token context, effective batch 16 (micro-batch 1, accumulation 16), and 207 optimizer steps. Training takes 4h 43m to 5h 34m on one 16 GB GPU (Section 5.5).

Validation selection included larger and newer candidates. Hybrid linear-attention models exceeded the card at 6,144 tokens: 22.5 GB for 0.8B and 23.9 GB for 2B, although the 0.8B model fits in 11.7 GB at 3,072. At the required context, the dense 1.2B model was the viable option.

\subsection{A deliberate train and inference asymmetry}

Training inputs are annotated spans (median 1,716 words); inference inputs are locator spans, up to 2,000 words at 0.670 utterance recall for promoted and median 2,712 words at 0.627 for protocol-exact, both on validation. Training data are therefore cleaner and shorter.

On the replaced 1.7B summarizer at the protocol budget, training instead on retrieved spans moved validation ROUGE-1 by \textbf{-0.51 [-1.68, +0.64]} (32.93 versus 33.44). Matching inference bought no measurable gain and introduced locator errors as training noise.

\section{Experimental setup}

We score complete splits under Section 3's protocol. Baselines span four families.

\subsection{Zero-shot proprietary hosted models}

We run GPT-5.6 Sol and Luna, plus Claude Opus 5, Sonnet 5 and Haiku 4.5. Each receives the full transcript, the Appendix A instruction, and disabled reasoning or thinking where supported.

Both vendors reject \texttt{temperature=0} on these models. GPT-5.6 must also disable reasoning or it can spend the 512-token budget reasoning and return an empty summary; Claude uses a separate thinking switch. We therefore use vendor-default sampling with \texttt{reasoning\_effort=none} for GPT-5.6 and thinking disabled for Claude. Released records contain the exact request parameters, enforced by tests.

Gemini 3.x is excluded because it rejects a zero thinking budget and cannot match this non-reasoning regime.

Exact reproduction under current vendor APIs is unavailable. Even at temperature 0, two Claude Haiku 4.5 passes over 30 queries agree exactly on only 13.3\% of predictions (the scored row uses vendor defaults). All five rows were collected on 27 July 2026 UTC. Responses expose floating aliases rather than dated snapshots, so we release the returned aliases, request times and outputs as provenance. Our local pipeline reproduces all 281 test predictions byte-identically.

Generating all five 281-query hosted-model rows through batch APIs cost \$37.33 in total, compared with \$74.65 at the vendors' synchronous rates. On 30-query controls, batch-versus-sync agreement matches or exceeds synchronous-versus-synchronous agreement for both tested models. Batched rows have no per-query latency; Section 5.5 measures latency separately.

The 512-token completion cap binds asymmetrically and against the verbose systems: over 281 queries it binds 116 times for Claude Opus 5, 140 for Claude Sonnet 5, three for Claude Haiku 4.5, eight for GPT-5.6 Sol and two for GPT-5.6 Luna. It binds zero times for our promoted configuration and once for the protocol-exact one. Section 6.1 therefore treats the hosted-model ordering as a reference-overlap result under this prompt and cap, not as a length-controlled quality comparison.

\subsection{Released community checkpoints}

We run four \texttt{mikeadimech} QMSum fine-tunes: distilbart, bart-large-cnn, pegasus and led-base.

Three have a 700-word input limit, confounding task-specific training with truncation. Led-base is degenerate under vanilla decoding and is reported without repair. The set is not exhaustive: we did not run other released BART-large, Flan-T5-large, DialogLED-base, Llama-2 or community LongT5-XL fine-tunes (no official LongT5 fine-tune was released). Claims cover only the checkpoints run.

These rows use \texttt{num\_beams=4}; ours is greedy. Four beams require 27.3 GB on our 1.2B model and do not fit the 16 GB card (Section 5.5).

A protocol-exact 2x2 over two-beam search and trigram repetition blocking finds no detectable ROUGE-1 gain, although summaries change (median per-query absolute change 4.4). Repetition blocking costs \textbf{1.5 ROUGE-2} at both beam settings: -1.62 [-2.28, -1.00] greedily and -1.44 [-2.18, -0.71] at two beams. We retain greedy decoding without blocking.

\subsection{A specialist we fine-tuned ourselves}

DialogLED-large, fine-tuned on identical targets and locator outputs for five epochs with beam 4, reaches 30.67 validation ROUGE-1. The same targets give the 1.7B instruct model 33.44, a 2.8-point deficit [+1.60, +3.93]. Its authors report 34.50 for a full-input fine-tune, so the span pipeline explains part of the gap. We do not spend a test touch on this baseline.

\subsection{The strongest publicly released specialist, via its own predictions}

\citet{pagnoni2023socratic} release their fine-tuned Segment Encoder, validation and test predictions, and scores for those files. We rescore the exact outputs; no inference is run for this row.

The files lack query IDs, so we recover monotone alignment by token similarity and gate it on the authors' scores; a one-position error would otherwise yield plausible low-twenties ROUGE. Their paper prints 38.06, but the released-file scores are 38.48 test and 38.53 validation, the relevant calibration targets. Our scorer returns 38.5982 and 38.5854. We found no released predictions for QontSum or LTRSum, hence ``strongest with released predictions.''

Released predictions avoid our port's offset (Sections 7.5 and 8.3). Other reruns are unavailable: Vig et al.'s hosting bucket denies anonymous access and listing (checked 2026-07-30), and one of Pagnoni et al.'s two Hub repositories contains only a README.

\subsection{Hardware, latency and cost}

Local training and inference use one NVIDIA RTX 5070 Ti (16 GB), Ryzen 7800X3D, Windows and PyTorch cu128.

On the two longest protocol-exact validation prompts (4,522 and 4,516 tokens), peak allocated memory is 8.76 GB greedily, 14.95 GB at two beams and 27.33 GB at four, increasing by \texttt{(beams - 1) x 6.19 GB} because prefill duplicates the prompt. \textbf{Four beams do not fit} (promoted prompts are about 30\% shorter and unprobed); host spill takes 430 s/query versus the promoted greedy mean of 2.35 s. Two beams retain 1.35 GB headroom.

Latency uses an idle device, CUDA synchronization and 280 timed test queries after warmup. Protocol-exact takes 0.022 s to locate, 2.607 s to generate and 2.629 s end to end at 29.1 decoded tokens/s. Promoted takes 0.071 s to locate and 2.42 s end to end on test, and 2.35 s and 2.76 s in the two validation runs of Section 7.5.

Training takes \textbf{4.7 to 5.6 GPU-hours}: \textbf{4h 43m} and \textbf{5h 34m} for the two completed runs, the second launched on an idle GPU. Peak allocated memory is 12.1 GB, but peak \textbf{reserved} is \textbf{30.9 GB on a 17.1 GB card} (16 GiB in allocator decimal GB), causing host-memory fallback. Once settled, 10.3 to 10.7 s/step would imply about 36 minutes over 207 steps, but no run avoids fallback long enough to realize that rate.

Locator training takes \textbf{about 7 minutes} and 5.27 GB for promoted, versus \textbf{about 1.5 minutes} and 2.82 GB for protocol-exact, roughly 2\% of training time.

Of four additional seeds, one completed in 5h 34m; one exhausted memory at step 2, one reached only step 40 after eight hours, and one degraded after step 45 (Section 8.9). The setup is therefore at the 16 GB card's edge. GPT-5.6 Luna takes 2.34 s synchronously over 30 queries, within our local pipeline's roughly 20\% timing noise. API latency varies 2.6x to 6.9x within one run and measures service rather than model speed, so we use it only for deployment context.

\section{Results}

\subsection{A single-protocol scale}

\label{sec:scale}

Table 1 uses the full test split (n=281) and one scorer. We generated all predictions except the Socratic-SegEnc released-output row, which uses the authors' outputs (Sections 5.4 and 6.3).

\begin{table}[H]
\centering
\small
\setlength{\tabcolsep}{5.5pt}
\caption{The single-protocol scale. Full official test split (n=281), one frozen scorer, at most one test touch per system. Every row generated and scored by us, except Socratic-SegEnc (released outputs), scored from the authors' predictions (Section 5.4).}
\label{tab:scale}
\begin{tabular}{>{\raggedright\arraybackslash}p{0.297\linewidth}lrrrrr}
\toprule
System & Params & R1 & R2 & R-L & R-Lsum & BERTScore \\
\midrule
Socratic-SegEnc (released outputs) & 406M & 38.60 & 13.91 & 24.98 & 33.71 & 0.8737 \\
Span-trained SegEnc & 406M & 36.33 & 12.72 & 23.69 & 32.17 & 0.8710 \\
Ours (promoted) & 1.2B + 33M & 35.41 & 12.28 & 24.63 & 31.36 & 0.8733 \\
Socratic-SegEnc (our port) & 406M & 35.30 & 11.87 & 23.00 & 30.56 & 0.8695 \\
Ours (protocol-exact) & 1.2B + 22.7M & 33.39 & 10.65 & 22.83 & 29.30 & 0.8680 \\
GPT-5.6 Luna (zero-shot) & undisclosed & 32.43 & 7.89 & 19.05 & 27.49 & 0.8608 \\
GPT-5.6 Sol (zero-shot) & undisclosed & 30.92 & 6.94 & 18.33 & 26.10 & 0.8583 \\
LFM2.5-1.2B (zero-shot, spans) & 1.2B + 33M & 30.12 & 6.70 & 19.29 & 25.04 & 0.8687 \\
Claude Opus 5 (zero-shot) & undisclosed & 28.87 & 8.43 & 16.97 & 25.01 & 0.8522 \\
Claude Haiku 4.5 (zero-shot) & undisclosed & 28.66 & 8.42 & 17.26 & 24.19 & 0.8431 \\
DistilBART & 306M & 28.65 & 6.54 & 17.92 & 25.50 & 0.8546 \\
LFM2.5-1.2B (zero-shot, truncated) & 1.2B & 28.57 & 5.55 & 17.79 & 24.55 & 0.8607 \\
Claude Sonnet 5 (zero-shot) & undisclosed & 27.89 & 7.53 & 16.11 & 23.98 & 0.8478 \\
BART-large-CNN & 406M & 27.32 & 5.65 & 17.41 & 24.08 & 0.8513 \\
PEGASUS & 570M & 20.09 & 4.45 & 14.92 & 17.23 & 0.8344 \\
LED-base & 162M & 9.41 & 2.37 & 7.95 & 7.96 & 0.7818 \\
\bottomrule
\end{tabular}
\end{table}

\artifactnote{Socratic-SegEnc appears as the authors' released outputs and as our inference port; the latter is a calibration, not a reproduction (Section 8.3). Hosted zero-shot rows use the full transcript. All BERTScores use the same scorer. Parameter counts are measured (\texttt{num\_parameters()}); the community checkpoints were fine-tuned by \texttt{mikeadimech}.}

\begin{figure}[t]
\centering
\includegraphics[width=\linewidth]{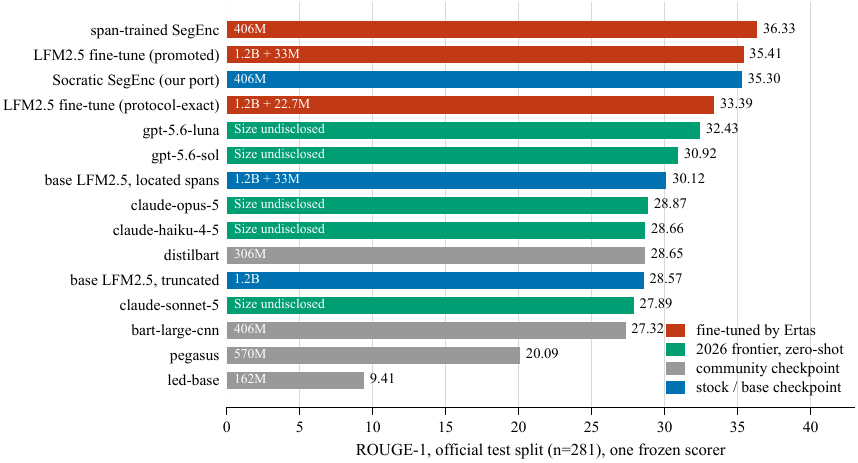}
\caption{The runnable-protocol scale. Every bar uses predictions generated through our evaluation path on the official test split (n=281) and one frozen scorer. The Socratic-SegEnc bar is our 35.30 port result, matching the separately labelled port row in Table~\ref{tab:scale}. Systems are sorted by ROUGE-1, with parameter counts written on the bars.}
\label{figure-scale}
\end{figure}

Read the table in this order: training scale, our system's position, the port calibration, then the fixed-model control. The released-output score remains the primary Socratic-SegEnc row because it uses the authors' pipeline; the adjacent 35.30 row shows the same checkpoint through our port for a stricter generated-output comparison. Fine-tuning that checkpoint for our span regime raises the port result to 36.33 on test. Sections 8.2 and 8.3 isolate why the two stock scores differ and why we do not transfer the 3.30-point port offset to the trained arm. Figure~\ref{figure-scale} shows this generated-output comparison.

\textbf{Under this prompt and scorer, the specialist leads the hosted-model rows.} A 406M Fusion-in-Decoder specialist from 2023 reaches 38.60 ROUGE-1, and the strongest of the five proprietary hosted models reaches 32.43. That is a gap of \textbf{6.2 ROUGE-1}, with a query-level 95\% interval of [+5.05, +7.31] and a meeting-cluster interval of [+4.84, +7.45]. On ROUGE-2 the same comparison is 13.91 against 7.89, a \textbf{76\% relative} difference; against the strongest hosted-model ROUGE-2, claude-opus-5's 8.43, the relative margin is still 65\%. Every proprietary hosted model tested, given the full transcript and its non-reasoning regime, scores below this task-trained checkpoint on these reference-overlap metrics.

\textbf{Length contributes to the F-measure gap.} References have a median of 59 words, while hosted answers are longer. Every hosted-model row has higher ROUGE-1 recall than ours and lower precision:

\begin{table}[H]
\centering
\small
\setlength{\tabcolsep}{5.5pt}
\begin{tabular}{lrrrrr}
\toprule
system, test split & R1 precision & R1 recall & R1 F & median words & x reference \\
\midrule
ours, protocol-exact & 0.359 & 0.348 & 0.334 & 56 & 0.95 \\
gpt-5.6-luna & 0.288 & 0.444 & 0.324 & 86 & 1.46 \\
gpt-5.6-sol & 0.279 & 0.417 & 0.309 & 81 & 1.37 \\
claude-opus-5 & 0.195 & 0.629 & 0.289 & 223 & 3.78 \\
claude-haiku-4-5 & 0.200 & 0.574 & 0.287 & 190 & 3.22 \\
claude-sonnet-5 & 0.192 & 0.592 & 0.279 & 214 & 3.63 \\
\bottomrule
\end{tabular}
\end{table}

\artifactnote{The decomposition run used the protocol-exact configuration; the promoted row was not decomposed.}

Claude outputs are 3.2x to 3.8x reference length, so the ordering partly reflects reference-anchored F-measures applied to systems not tuned to QMSum length. The shared prompt has no length hint (Appendix A); Section 8.1 scopes the claim.

\textbf{Our locate-then-summarize system falls between the groups.} It reaches \textbf{35.41 ROUGE-1 and 0.873 BERTScore} at 1.2B parameters plus a 33M locator, trained in about five GPU-hours on one 16 GB consumer GPU, above all five hosted-model baselines on ROUGE-1 and 3.2 ROUGE-1 below the specialist. Section 7.5 retrains that specialist for our regime; the resulting test difference is unresolved while measured resource use is lower.

Every margin against our system carries a paired bootstrap interval over the 281 test queries, 10,000 resamples, resampling queries jointly so that between-query variance cancels:

\begin{table}[H]
\centering
\small
\setlength{\tabcolsep}{5.5pt}
\begin{tabular}{>{\raggedright\arraybackslash}p{0.320\linewidth}rl}
\toprule
comparison & delta ROUGE-1 & 95\% interval \\
\midrule
vs claude-sonnet-5 & +7.52 & [+6.24, +8.79] \\
vs distilbart (best community checkpoint) & +6.76 & [+5.24, +8.30] \\
vs claude-haiku-4-5 & +6.74 & [+5.49, +7.97] \\
vs claude-opus-5 & +6.54 & [+5.23, +7.80] \\
vs gpt-5.6-sol & +4.49 & [+3.24, +5.77] \\
vs gpt-5.6-luna & +2.98 & [+1.75, +4.24] \\
\textbf{vs stock Socratic SegEnc (authors' predictions)} & \textbf{-3.19} & \textbf{[-4.51, -1.83]} \\
\bottomrule
\end{tabular}
\end{table}

\begin{figure}[t]
\centering
\includegraphics[width=\linewidth]{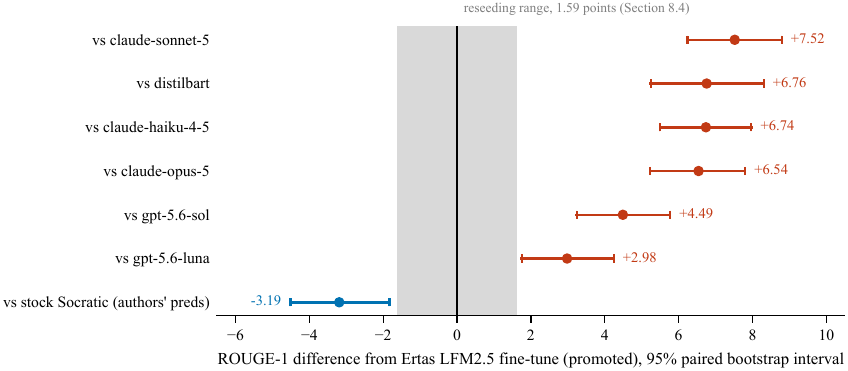}
\caption{Every margin against our promoted system, with both sources of uncertainty. Bars are 95\% paired bootstrap intervals over the 281 test queries. The shaded band is the 1.59-point reseeding range of Section~\ref{sec:seed-variance}, which the paired bootstrap does not capture.}
\label{figure-margins}
\end{figure}

All seven query-level intervals exclude zero. We treat this family descriptively rather than as seven independent confirmatory tests and apply no familywise correction. Section 3.4 separately reports meeting-cluster intervals for the three headline comparisons; their conclusions are unchanged. Figure~\ref{figure-margins} shows the seven margins against the reseeding range of Section~\ref{sec:seed-variance}.

Bootstrap intervals do not capture training-run variance. Reseeding ours moves ROUGE-1 by \textbf{1.59} (Section 8.4). Every winning margin exceeds that range, but +2.98 against gpt-5.6-luna is under twice it, and the 3.19-point specialist deficit is not immune to it.

\textbf{The distance from our own base model is chiefly the fine-tune.} Table 1's two base-model rows are the promoted configuration's ablation cells (Section 7.4): the same LFM2.5-1.2B with no adapter, zero-shot under its chat template, reading either the same located spans or a 4,500-word truncated transcript. At identical retrieval the fine-tune is worth \textbf{+5.29 ROUGE-1, 95\% interval [+4.02, +6.56]}, and nearly doubles ROUGE-2 (from 6.70 to 12.28), while replacing truncation with located spans without fine-tuning moves ROUGE-1 by +1.55 [+0.58, +2.51] on test and +0.29 on validation. Training is the larger measured lever.

\subsection{Two configurations, and why we report both}

The promoted row changes the frozen locator window from 900 to 375 words and span budget from 3,000 to 2,000 (Section 3.2), so Table 1 includes both configurations:

\begin{table}[H]
\centering
\small
\setlength{\tabcolsep}{5.5pt}
\begin{tabular}{>{\raggedright\arraybackslash}p{0.320\linewidth}rrrrr}
\toprule
configuration (test split) & R1 & R2 & R-L & R-Lsum & BERTScore \\
\midrule
protocol-exact & 33.39 & 10.65 & 22.83 & 29.30 & 0.868 \\
promoted (the deviating configuration) & \textbf{35.41} & \textbf{12.28} & \textbf{24.63} & \textbf{31.36} & \textbf{0.873} \\
\bottomrule
\end{tabular}
\end{table}

The difference is \textbf{+2.01 ROUGE-1, 95\% interval [+0.77, +3.26]}.

Protocol-exact scores 33.89 on validation and 33.39 on test; promoted scores 35.39 and 35.41. Selection used five validation candidates and no test tuning.

\textbf{The margin does not survive reseeding.} With a different summarizer seed, promoted scores \textbf{33.80 validation}, below protocol-exact's 33.89. The 1.59-point seed range matches the original 1.5-point validation gain, so we cannot show configuration robustness (Section 8.4).

\subsection{What the scorer contributes, measured rather than assumed}

The specialist's released predictions and scores let us measure the scorer offset.

Across 553 predictions, our scorer gives \textbf{38.5982} versus their \textbf{38.48} on test and \textbf{38.5854} versus \textbf{38.53} on validation: differences of \textbf{+0.12 and +0.06 ROUGE-1}. Their paper prints 38.06; Section 5.4 explains why released-file scores are the calibration targets.

Thus scorer choice cannot explain our 3.2-point deficit, and the published 37 to 39 band is roughly commensurable with our scale. This one calibration does not license merging other published numbers, whose protocols and even test counts differ (279 reported versus 281 released).

\subsection{An embedding metric barely separates these systems}

For the specialist, ours and the strongest hosted-model baseline, BERTScore is 0.874 / 0.873 / 0.861, whereas ROUGE-2 is 13.91 / 12.28 / 7.89. The embedding metric compresses the separation.

For every baseline we beat on ROUGE-1, we also lead on BERTScore with a paired interval excluding zero:

\begin{table}[H]
\centering
\small
\setlength{\tabcolsep}{5.5pt}
\begin{tabular}{>{\raggedright\arraybackslash}p{0.320\linewidth}rl}
\toprule
comparison (ours minus theirs) & delta BERTScore & 95\% interval \\
\midrule
vs led-base & +0.0915 & [+0.0866, +0.0964] \\
vs pegasus & +0.0389 & [+0.0352, +0.0424] \\
vs claude-haiku-4-5 & +0.0302 & [+0.0280, +0.0323] \\
vs claude-sonnet-5 & +0.0255 & [+0.0232, +0.0277] \\
vs bart-large-cnn & +0.0220 & [+0.0190, +0.0249] \\
vs claude-opus-5 & +0.0211 & [+0.0189, +0.0233] \\
vs distilbart & +0.0187 & [+0.0158, +0.0216] \\
vs gpt-5.6-sol & +0.0150 & [+0.0127, +0.0173] \\
vs gpt-5.6-luna & +0.0125 & [+0.0102, +0.0148] \\
\textbf{vs stock Socratic SegEnc (authors' predictions)} & \textbf{-0.0005} & \textbf{[-0.0029, +0.0020]} \\
\bottomrule
\end{tabular}
\end{table}

The result holds against every hosted-model and community checkpoint we outscore on ROUGE-1; no BERTScore resample crosses zero (R=10,000). BERTScore also reorders the middle of Table 1: distilbart exceeds claude-opus-5, and bart-large-cnn exceeds claude-sonnet-5 and claude-haiku-4-5, despite trailing them on ROUGE-1.

The specialist, however, beats us by 3.2 ROUGE-1 with an interval firmly excluding zero, and on BERTScore the same pair of systems is statistically level. One metric calls that comparison decisive and the other cannot separate them at all. The same pattern appears in Section 7.5, where ROUGE and BERTScore point in opposite directions on an insignificant comparison. We therefore lead with ROUGE-1 and ROUGE-2.

\section{Analysis and ablations}

All ablations are on validation. Several correct our earlier conclusions; in three cases, the error is more informative than the original conclusion.

\subsection{Retrieval window size should be set by the encoder's context}

Our locator scored 900-word windows with a cross-encoder whose context is 512 tokens. A 900-word window of speaker-labelled transcript is a median of 1,150 tokens, so \textbf{97.3\% of windows exceeded the encoder's context and the ranking cross-encoder received a mean of 47.7\% of the text it was ranking.} The silent truncation can hide relevant text.

Setting the window to 375 words, so that a window fits the encoder whole (utterance-level recall on validation):

\begin{table}[H]
\centering
\small
\setlength{\tabcolsep}{5.5pt}
\begin{tabular}{lrrr}
\toprule
locator & window & utterance recall @2000w & @3000w \\
\midrule
MiniLM-L6 (original) & 900 & 0.5035 & 0.6268 \\
MiniLM-L12 & 900 & 0.5108 & 0.6319 \\
MiniLM-L6 & 375 & 0.6420 & 0.7468 \\
\textbf{MiniLM-L12} & \textbf{375} & \textbf{0.6704} & \textbf{0.7650} \\
\bottomrule
\end{tabular}
\end{table}

Doubling encoder depth at the wrong window size is worth \textbf{+0.005} recall. Fixing the window size is worth \textbf{+0.120}, a factor of 24 at the 3,000-word budget (about 19 at 2,000). Narrowing the window also rebuilds the ranking cross-encoder's training pairs (13,738 to 33,655), so the intervention is window-plus-retraining rather than window alone. The interaction confirms the mechanism: the deeper encoder gains 0.005 when it sees 47.7\% of a window and 0.018 when it sees all of it, so capacity only pays once the input contains the evidence.

As Section 4.1 defines, utterance recall is comparable across window widths; window recall is not.

\subsection{Retrieval recall and summarization quality are decoupled}

Raising utterance recall from 0.627 to 0.765 at a fixed 3,000-word budget moved ROUGE-1 by \textbf{+0.61, 95\% interval [-0.49, +1.69]}, inside noise. The promoted configuration instead cuts the budget to 2,000 words and accepts recall 0.670. At the fixed new locator that cut alone is worth \textbf{+0.89, [-0.10, +1.86]}, on validation (34.50 to 35.39), an interval that crosses zero on its own; the combined change over the protocol-exact configuration is \textbf{+1.49, [+0.42, +2.56]} on ROUGE-1, with the same picture on ROUGE-2 (+1.26, [+0.32, +2.23]) and ROUGE-Lsum (+1.19, [+0.18, +2.19]). One reseeding erases the combined gain (Section 8.4), so the durable result is a non-monotone relationship between recall and ROUGE, not the exact optimum.

The measured curve has an interior maximum:

\begin{table}[H]
\centering
\small
\setlength{\tabcolsep}{5.5pt}
\begin{tabular}{rrr}
\toprule
span budget & utterance recall & val ROUGE-1 \\
\midrule
1,500 & 0.610 & 34.70 \\
\textbf{2,000} & \textbf{0.670} & \textbf{35.39} \\
3,000 & 0.765 & 34.50 \\
\bottomrule
\end{tabular}
\end{table}

Gold-span prompts have a median of 1,680 mostly relevant words; 3,000-word retrieved prompts contain about 500 annotated and 2,500 other words. This evidence suggests evaluating span precision and downstream utility alongside recall.

Unlike a withdrawn earlier analysis confounded by train/inference length mismatch, this comparison holds window size, locator training and summarizer fixed and varies only the budget.

Linear scaling from recall predicted +1.24 ROUGE-1 but measured +0.61, so we reject that extrapolation.

\subsection{Perfect retrieval does not reach the specialist, and retrieval is the smaller half}

On the 237 specific validation queries, with the same summarizer throughout:

\begin{table}[H]
\centering
\small
\setlength{\tabcolsep}{5.5pt}
\begin{tabular}{>{\raggedright\arraybackslash}p{0.320\linewidth}r}
\toprule
configuration & R1 \\
\midrule
our pipeline & 35.61 \\
our gold-span oracle & 36.71 \\
stock Socratic SegEnc (authors' predictions) & 38.10 \\
\bottomrule
\end{tabular}
\end{table}

The retrieval lever is \textbf{+1.09, 95\% interval [-0.22, +2.43]}, and the residual from a perfect locator to the specialist is \textbf{+1.40, [+0.05, +2.76]}. The lever's interval crosses zero and the residual's excludes it. Retrieval accounts for 44\% of the 2.49-point gap as a point estimate and the summarizer for 56\%; +1.09 is close to this benchmark's paired detection floor of roughly 0.9 to 1.4 ROUGE-1. \textbf{A perfect locator is worth about what this benchmark can resolve.}

Before the window fix, the lever was +3.35 and the residual +1.23. The corrected promoted configuration does not support treating retrieval as the bottleneck; these superseded values explain the reversed conclusion.

One further comparison from the same rows: our 1.2B summarizer and the 1.7B model it replaced score 36.71 and 36.87 given gold spans, a difference of +0.16 [-1.07, +1.43]. \textbf{Given clean input the benchmark does not statistically separate the two backbones}, so the base-model axis that appeared to win our model-selection sweep was largely the locator.

\subsection{Decomposition beats naive truncation; tested decoding settings do not help}

\label{sec:levers}

Fine-tuning Qwen3-1.7B, the instruct model our promoted summarizer replaced, on truncated full transcripts rather than on annotated spans scores 22.72 against 33.44, a gap of 10.7 ROUGE-1 [+9.05, +12.38], at roughly an order of magnitude more compute per step. The truncation arm stopped after one of three epochs because of per-step cost, so it is undertrained, and first-4,500-word truncation can omit later evidence. This compares decomposition with naive truncation, not with a model that can hold the full transcript.

Fixed-model controls remove these confounds: the same zero-shot LFM2.5-1.2B reads either located spans or the 4,500-word truncation, and the promoted row adds only fine-tuning.

\begin{table}[H]
\centering
\small
\setlength{\tabcolsep}{5.5pt}
\begin{tabular}{>{\raggedright\arraybackslash}p{0.320\linewidth}rrrrr}
\toprule
cell, official test split & R1 & R2 & R-L & R-Lsum & BERTScore \\
\midrule
base, truncated transcript & 28.57 & 5.55 & 17.79 & 24.55 & 0.8607 \\
base, our located spans & 30.12 & 6.70 & 19.29 & 25.04 & 0.8687 \\
fine-tuned, our located spans (promoted) & 35.41 & 12.28 & 24.63 & 31.36 & 0.8733 \\
\bottomrule
\end{tabular}
\end{table}

\begin{figure}[t]
\centering
\includegraphics[width=\linewidth]{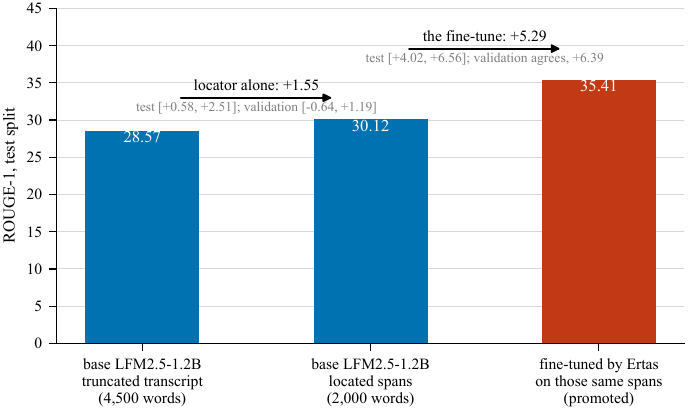}
\caption{The two levers, de-confounded. One base model, one adapter, three conditions: only the input regime changes between the first two bars, and only the fine-tune between the second and third.}
\label{figure-levers}
\end{figure}

\textbf{The fine-tune is the lever: +5.29 ROUGE-1, 95\% interval [+4.02, +6.56], at identical retrieval} (validation agrees, +6.39 [+5.06, +7.69]). Both zero-shot cells produce near-reference lengths (median 45 and 65 words versus 59) with no empty generations. Figure~\ref{figure-levers} separates the two levers across the three conditions.

\textbf{The locator alone is worth little to the untuned model, and the two splits disagree on whether it is worth anything: +1.55 [+0.58, +2.51] on test against +0.29 [-0.64, +1.19] on validation.} Together these are a smaller effect than fine-tuning. The locator also gives 1.8x lower latency and 2.9x lower peak memory (5.70 against 16.57 GB, validation runs), because 2,000 located words are well under half the tokens of the truncation (median 2,581 against 5,833).

Together with Section 7.5, the specialist loses 6.3 off-regime, replacing the first 4,500 words with 2,000 retrieved words moves the base model by +1.55 on test and +0.29 on validation, and target-regime training adds 5.3 to 6.4. Training is the larger measured lever.

Across the settings tested in Section 5.2, beam search and trigram blocking do not improve ROUGE-1; 140 queries improve and 132 worsen, median absolute change 4.4. Blocking costs 1.5 ROUGE-2, and four beams do not fit the card (Section 5.5), so we decode greedily.

\subsection{The specialist's advantage follows its training regime}

\label{sec:regime}

Through our port, validation decomposes the stock checkpoint's regime transfer and its recovery:

\begin{table}[H]
\centering
\small
\setlength{\tabcolsep}{5.5pt}
\begin{tabular}{>{\raggedright\arraybackslash}p{0.267\linewidth}>{\raggedright\arraybackslash}p{0.267\linewidth}r>{\raggedright\arraybackslash}p{0.267\linewidth}}
\toprule
checkpoint & source regime & val R1 & change \\
\midrule
stock Socratic SegEnc & capped long input, about 11,800 words & 35.30 & baseline \\
stock Socratic SegEnc & our located spans, 2,000 words & \textbf{29.00} & \textbf{-6.30 vs stock long-input} \\
span-trained SegEnc, reported checkpoint & the same located spans & \textbf{37.05} & \textbf{+8.05 vs stock-on-spans; +1.75 net} \\
\bottomrule
\end{tabular}
\end{table}

The first two rows hold weights and padded chunk count fixed, so \textbf{-6.30, 95\% interval [-7.51, -5.12]}, is the within-model regime drop. The +8.05 recovery is the directly observed point difference from adapting the checkpoint to the source text used at inference; the net +1.75 over stock long-input carries a paired interval of \textbf{[+0.54, +2.93]}. At 29.00, the off-regime checkpoint is separately 6.39 below our 35.39 validation row. Section 8.3 treats the port's 3.3-point absolute offset from the authors' released predictions separately.

Read alone, the table suggests Fusion-in-Decoder requires long input. That conclusion is unsupported: input and training regime vary together.

We fine-tune the checkpoint on span-regime source text \textbf{verified byte-identical} to our summarizer's. At 63 chunks, validation ROUGE-1 rises from 29.00 to 35.18 after one epoch, then 36.57, 35.83 and 37.49.

The reported 15-chunk fine-tune, matched to 2,000-word spans, reaches \textbf{37.13 at epoch 4 versus 37.49} for 63 chunks, a 0.36 difference [-0.44, +1.14], in 22 rather than 86 minutes. Four lower-learning-rate continuation epochs give:

\begin{table}[H]
\centering
\small
\setlength{\tabcolsep}{5.5pt}
\begin{tabular}{rr>{\raggedright\arraybackslash}p{0.320\linewidth}}
\toprule
effective epoch, matched-chunk lineage & val R1 &  \\
\midrule
4 & 37.13 & end of the initial fine-tune, 22 minutes \\
5 & 38.31 & peak \\
6 & 36.59 &  \\
7 & 36.97 &  \\
8 & \textbf{37.05} & last checkpoint, the figure we report \\
\bottomrule
\end{tabular}
\end{table}

\begin{figure}[t]
\centering
\includegraphics[width=\linewidth]{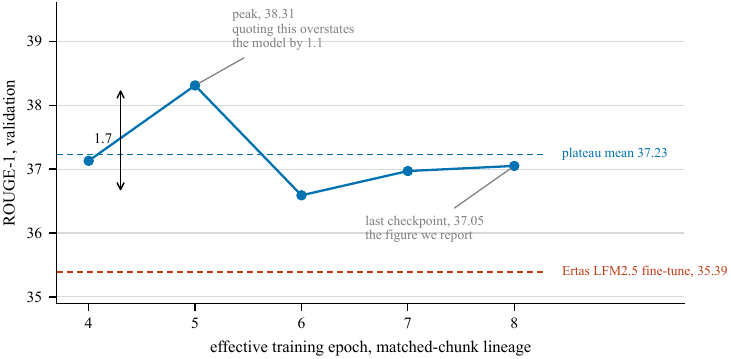}
\caption{What one model's training epochs are worth on this benchmark. Five checkpoints of the span-trained specialist differing only in training epoch, on validation. Quoting the peak rather than the plateau would overstate the model by 1.1 ROUGE-1.}
\label{figure-epochs}
\end{figure}

\textbf{We report the plateau, not the peak.} Epochs 5 to 8 average 37.23 (standard deviation 0.65); epoch 5 is 1.44 above the next three's mean. Quoting 38.31 would overstate the model by 1.1 over the plateau and 1.3 over our reported checkpoint. The four-epoch continuation length was chosen after the first four epochs, so checkpoint selection is absent but stopping remains a choice. Figure~\ref{figure-epochs} plots all five checkpoints.

Against our promoted configuration, 35.39 on the same validation queries:

\begin{table}[H]
\centering
\small
\setlength{\tabcolsep}{5.5pt}
\begin{tabular}{>{\raggedright\arraybackslash}p{0.320\linewidth}rl}
\toprule
basis & R1 & margin over our system \\
\midrule
\textbf{last checkpoint, without checkpoint selection} & \textbf{37.05} & \textbf{+1.67, [+0.58, +2.76]} \\
plateau mean, epochs 5 to 8 & 37.23 & about +1.8 \\
worst checkpoint after the peak & 36.59 & +1.20 \\
peak, reported for completeness only & 38.31 & +2.93, [+1.88, +3.99] \\
\bottomrule
\end{tabular}
\end{table}

\artifactnote{The plateau-mean and worst-checkpoint rows are curve summaries rather than sampled margins; the bracketed rows carry the intervals.}

Every checkpoint from epoch 4 onward leads ours on validation by 1.2 to 2.9 ROUGE-1; 37.23 is their plateau mean, while 35.39 is our maximum over five candidates.

\textbf{The validation margin does not replicate on test.} We take the reported checkpoint to the test split once:

\begin{table}[H]
\centering
\small
\setlength{\tabcolsep}{5.5pt}
\begin{tabular}{lrrrl}
\toprule
metric & ours & span-trained SegEnc & diff & 95\% interval \\
\midrule
ROUGE-1 & 35.41 & \textbf{36.33} & +0.93 & \textbf{[-0.42, +2.24]} \\
ROUGE-2 & 12.28 & \textbf{12.72} & +0.43 & [-0.77, +1.64] \\
ROUGE-L & \textbf{24.63} & 23.69 & -0.94 & [-2.06, +0.18] \\
ROUGE-Lsum & 31.36 & \textbf{32.17} & +0.81 & [-0.47, +2.05] \\
BERTScore & \textbf{0.8733} & 0.8710 & -0.0022 & [-0.0046, +0.0001] \\
\bottomrule
\end{tabular}
\end{table}

\textbf{Every interval crosses zero.} ROUGE-1, ROUGE-2 and ROUGE-Lsum point toward the 406M model, while ROUGE-L and BERTScore point toward ours. The +1.67 validation margin becomes +0.93 on test. The meeting-cluster ROUGE-1 interval is [-0.27, +2.22], so accounting for queries nested within meetings does not change the conclusion.

The checkpoint was fixed before test, so the split discrepancy illustrates the detection-floor effects in Section 3.4 rather than test selection.

The test split does not statistically separate the systems; measured resource use differs:

\begin{table}[H]
\centering
\small
\setlength{\tabcolsep}{5.5pt}
\caption{Resource comparison at an unresolved test difference. Both systems include the shared locator; ratios divide our measured value by the span-trained SegEnc value. Crossing intervals do not establish formal equivalence.}
\label{tab:resources}
\begin{tabular}{>{\raggedright\arraybackslash}p{0.320\linewidth}>{\raggedright\arraybackslash}p{0.209\linewidth}ll}
\toprule
 & span-trained SegEnc & ours & ratio \\
\midrule
parameters, incl. the shared 33M locator & \textbf{439M} & 1.2B + 33M & \textbf{2.8x} \\
peak inference VRAM, max over the validation split & \textbf{2.66 GB} & 5.73 GB & \textbf{2.16x} \\
peak inference VRAM, median & \textbf{2.66 GB} & 4.71 GB & 1.77x \\
mean latency per query, two validation runs each, locate step included & \textbf{1.85s and 2.00s} & 2.35s and 2.76s & 1.3x to 1.4x \\
training to this checkpoint & \textbf{about 45 minutes} (eight matched-chunk epochs; the first four measured at 22 minutes) & 4.7 to 5.6 hours & large \\
test ROUGE-1 & 36.33 & 35.41 & difference unresolved \\
\bottomrule
\end{tabular}
\end{table}

Both sides include the shared 33M locator and 0.07 s locate step. Memory is summarizer peak allocated bytes with resident weights over the full validation split; the locator's 0.1 to 0.3 GB stage peak moves neither maximum.

The specialist uses 2.66 GB on all 272 validation queries because fixed-chunk padding fixes encoder shape. Ours ranges from 4.25 to 5.73 GB and tracks prompt length (r=0.999). Padding therefore trades compute for predictable memory.

Idle-GPU reruns produce \textbf{byte-identical predictions} while mean latency moves 17.2\% for ours and 8.5\% for theirs. The ordering survives every pairing, but fine-grained ratios do not.

At a test difference this benchmark cannot resolve, a 406M encoder-decoder reaches our operating point with about one-third the parameters, less than half the peak memory, lower measured latency, and about 45 minutes of training. This is non-rejection of a difference, not an equivalence result; we make no claim about unmeasured cost axes.

Within our port, span-regime training more than recovers the input-regime collapse. The span-trained model, reading 2,000 retrieved words, scores above the stock checkpoint reading about 11,800 in the capped long-input regime: \textbf{+1.75 [+0.54, +2.93]} at the reported checkpoint (and +3.01 [+1.96, +4.04] at the peak the plateau discipline rejects), against the long-input port's 35.30. Whether span training also beats the authors' own full-input pipeline is a question this experiment cannot decide: the span-trained arm was trained and evaluated inside our port, so it is native to the port's conventions, while the full-input arm carries the port's 3.3-point offset (Section 8.3), and against the authors' released full-input predictions the span-trained model sits 2.3 ROUGE-1 lower on test. The within-port statement is the one the evidence supports.

The published configuration's padding is waste at this input length. Fifteen chunks reach the same one-epoch training loss as 63 (1.7107 versus 1.7119) in \textbf{3.9x less time}. We report the shorter setting; 63 chunks were designed for long input, where the padding is load-bearing.

Neither system is guaranteed a whole meeting: the specialist caps its input at about 11,800 words, binding on the longest transcripts, we retrieve 2,000 on every query, and the median test transcript is 9,206 words, so the first table in this subsection is a roughly 6x comparison between capped regimes. Nothing here separates Socratic pretraining from BART-large's summarization prior; that would need a third condition we do not run. The regime comparison rests on within-port deltas, not absolute values.

\section{Limitations}

\subsection{What these metrics do not measure}

ROUGE and BERTScore do not measure whether individual propositions match the reference. In concurrent work, proprietary hosted baselines rank below ours on these metrics but above it under fact-level scoring; the community-checkpoint ordering holds under both. The 406M specialist was not included, so the specialist-versus-hosted-model result is established only on the metrics reported here. The shared prompt gives no target length, and the 512-token cap binds unevenly across hosted rows (Section 5.1). We did not run a length-controlled hosted-model comparison; the ordering is therefore limited to the submitted prompt, cap and scorer.

Section 6.4 shows the same limitation internally. The three systems above 35 ROUGE-1 span 3.19 ROUGE-1 but 0.0027 BERTScore; the specialist's BERTScore difference from ours is -0.0005 [-0.0029, +0.0020]. Automatic metrics disagree about whether this gap exists.

\subsection{The released-prediction row was not run by us}

Table 1's Socratic-SegEnc released-output row uses the authors' predictions, so its decoding and preprocessing cannot be ablated. Our scorer gives 38.5982 versus their 38.48 on test and 38.5854 versus 38.53 on validation, bounding only the scorer offset at 0.12 (Section 5.4). Figure 1 therefore does not use this row: its stock bar matches the separately labelled 35.30 port row in Table 1, so every bar in that chart comes from outputs generated through our evaluation path.

Our port lands 3.3 ROUGE-1 below those predictions (Section 8.3), and the remaining checkpoints in the line are unavailable (Section 5.4). The strongest row is therefore scorable but not end-to-end reproducible.

\subsection{Our port is not a reproduction}

Our Segment Encoder port scores 35.30 in the validation long-input regime versus 38.59 for released predictions, a 3.28-point offset. A preregistered test calibration gives 35.30 versus 38.60, an offset of 3.30. This offset includes preprocessing and decoding differences as well as any implementation difference; the 0.12 scorer calibration cannot explain it. We therefore use the port for Figure 1 and within-port deltas, while Table 1 presents the 35.30 port result beside the separately labelled 38.60 released-output result.

The span-trained arm is native to the port, so the offset cannot be transferred to it. We neither add 3.30 to its score nor describe 38.60 and 36.33 as a controlled training comparison. The controlled sequence is the one in Section 7.5: stock long-input 35.30, stock on our spans 29.00, and span-trained on those spans 37.05 on validation; on test, the port stock and span-trained rows score 35.30 and 36.33. These are checkpoint-level results from one training lineage, not an estimate over training seeds.

\subsection{Run-to-run variance}

\label{sec:seed-variance}

The checkpoint-choice measurement in Section 3.4 spans 1.7 ROUGE-1 within one run, comparable to the full-split detection floor of roughly $\pm$0.9 to $\pm$1.4. Cross-system margins involving our trained summarizer also depend on its training seed, so we measure that variation separately.

Holding all but seed fixed, \textbf{one of four further runs completed} (Section 8.9). The resulting dispersion estimate therefore contains \textbf{two runs}:

\begin{table}[H]
\centering
\small
\setlength{\tabcolsep}{5.5pt}
\begin{tabular}{lrrrr}
\toprule
seed & ROUGE-1 & ROUGE-2 & ROUGE-Lsum & BERTScore \\
\midrule
20260723 (promoted) & 35.39 & 11.39 & 31.09 & 0.8724 \\
101 & 33.80 & 10.37 & 29.86 & 0.8701 \\
\textbf{range} & \textbf{1.59} & \textbf{1.02} & \textbf{1.23} & \textbf{0.002} \\
\bottomrule
\end{tabular}
\end{table}

\textbf{Changing only seed moves ROUGE-1 by 1.59 points}, above the paired detection floor and near the 1.7-point checkpoint spread. Two runs provide a loose range, not a sampling distribution.

Fixed-checkpoint locator, decoding and oracle ablations are unaffected. Cross-system margins involving ours are qualified: the +2.98 to +7.52 hosted-model margins and 3.19-point specialist deficit exceed the range, although the smallest is under twice it. The +0.93 span-trained SegEnc comparison already crosses zero.

The 1.5-point promoted validation gain is not robust in this check: the reseeded promoted run scores \textbf{33.80}, below protocol-exact's \textbf{33.89}. One reseeding neither validates nor refutes the locator change; the measured gain applies to one training run.

The same seed change moves BERTScore by only 0.002, again showing its low sensitivity here.

The span-trained SegEnc uses one training lineage with multiple checkpoints but no independent training seeds. Its paired and meeting-cluster intervals condition on the reported checkpoint's predictions and do not include training-run variance. The resource comparison in Section 7.5 must therefore be read as a checkpoint-level comparison, not a population estimate over training runs.

\subsection{Neither system is guaranteed a whole meeting}

Section 7.5 compares capped regimes: the Segment Encoder accepts about 11,800 words, binding on the longest transcripts; ours uses 2,000; the median test transcript is 9,206 words. This is not a full-transcript-versus-retrieval comparison.

Located spans are off-distribution for the released checkpoint, so its 6.3-point drop measures transfer between regimes, not an architectural advantage for our summarizer.

Retrained on spans, it reaches 36.33 versus our 35.41, +0.93 [-0.42, +2.24], and the meeting-cluster interval is [-0.27, +2.22]. The benchmark does not separate the systems, which shows transfer without establishing equivalence. Socratic pretraining cannot be separated from BART-large's prior without another arm.

\subsection{The baseline checkpoints are uneven}

Three community checkpoints accept only 700 words, confounding task training with truncation; led-base is degenerate under vanilla decoding; and the four are not exhaustive. Claims cover the checkpoints run (Section 5.2).

They use four beams; ours is greedy. Tested settings do not improve our ROUGE-1, but four beams also require 27.3 GB and do not fit our 16 GB card.

\subsection{Two stages, trained separately}

Our stages are trained independently; DYLE shows that joint training can help on this task. Gold spans add 1.09 ROUGE-1 [-0.22, +2.43], and the oracle remains 1.40 below the specialist [+0.05, +2.76], bounding the headroom available to retrieval in our configuration.

\subsection{Scope}

One benchmark, one language, three meeting domains. Nothing establishes transfer to other query-focused summarization data. We report no human evaluation and only one operating point per configuration, not a cost-quality frontier.

\subsection{Training cost is measured on one machine, at the edge of its memory}

Section 5.5 gives the measurements. They come from one consumer GPU under Windows and do not generalize to other hardware. Host-memory fallback makes training slow and seed-dependent: only one of four additional seeds completes. Because failed seeds are dropped for a hardware reason, memory feasibility may select training orders in a way we cannot show is independent of final quality. Idle-device latency also varies by up to 17.2\%, so margins below about 20\% are measurement noise; peak memory, token counts and predictions reproduce exactly.

\subsection{Licence}

The LFM Open License v1.0 permits commercial use below US\$10M annual revenue and uncapped qualified non-profit and research use (verified 2026-08-07). This downstream restriction applies to our adapter, not the rest of the released artifacts.

\section{Conclusion}

Training a specialist on the input regime used at inference is the most robust result of this study. Through our port, a released 406M Segment Encoder loses 6.3 ROUGE-1 when moved from capped long input to 2,000 retrieved words, and fine-tuning on that span regime recovers the loss.

On test, the span-trained 406M model scores 36.33 ROUGE-1 and our 1.2B system scores 35.41. The test split does not statistically separate them on any reported metric; the meeting-cluster ROUGE-1 interval is [-0.27, +2.22]. The smaller model uses about one-third as many total parameters, less than half the peak inference memory, and about 45 minutes of training rather than roughly five hours. These are the measured resource axes; we do not claim formal equivalence or an unmeasured deployment advantage.

Fixed-model controls sharpen the attribution. Giving the zero-shot 1.2B base 2,000 retrieved words rather than the first 4,500 transcript words adds 1.55 ROUGE-1 on test and 0.29 on validation, whereas fine-tuning the same model for the span regime adds 5.29 [+4.02, +6.56]. Gold spans add 1.09 [-0.22, +2.43] on validation and leave a 1.40-point residual to the specialist [+0.05, +2.76]. Thus retrieval is useful for fitting the problem to local hardware, but fine-tuning supplies most of the measured quality gain.

Separately, under one concise prompt and reference-overlap scorer, the released 406M specialist exceeds five proprietary hosted models by at least 6.2 ROUGE-1. Their longer outputs and the absence of human and factuality evaluation prevent interpreting this as a general quality ordering.

Retrieval recall is also an incomplete target at this operating point: a 0.14 recall increase at a fixed budget yields no measurable ROUGE gain, while a shorter retrieved input scores best. This suggests evaluating precision and downstream utility alongside recall, not optimizing recall alone.

These conclusions are limited to ROUGE and BERTScore on one English benchmark. They do not establish human preference or factuality, and differences near one ROUGE-1 point approach QMSum's full-split detection floor. We therefore report paired intervals, preserve the labelled protocol deviation, and release the scorer, per-query predictions and trained artifacts needed to reproduce the scale.

\section*{Availability}

The Apache-2.0 repository \href{https://github.com/ErtasAI/qmsum-retrieved-span-training}{\texttt{https:/\allowbreak{}/\allowbreak{}github.\allowbreak{}com/\allowbreak{}ErtasAI/\allowbreak{}qmsum-\allowbreak{}retrieved-\allowbreak{}span-\allowbreak{}training}} contains the protocol, scorer, pipeline and per-query predictions. One command sequence reproduces our test row without API keys; proprietary hosted rows require vendor keys and are not exactly reproducible (Section 5.1). Data are rebuilt from QMSum's MIT-licensed release.

Trained artifacts are on the Hugging Face Hub at these pinned commits.

\begin{table}[H]
\centering
\small
\setlength{\tabcolsep}{5.5pt}
\begin{tabular}{>{\raggedright\arraybackslash}p{0.297\linewidth}>{\raggedright\arraybackslash}p{0.297\linewidth}>{\raggedright\arraybackslash}p{0.297\linewidth}}
\toprule
Artifact & Repository & Revision \\
\midrule
Summarizer adapter, promoted configuration & \href{https://huggingface.co/ErtasAI/qmsum-summarizer-lfm2.5-1.2b-lora}{\texttt{ErtasAI/\allowbreak{}qmsum-\allowbreak{}summarizer-\allowbreak{}lfm2.\allowbreak{}5-\allowbreak{}1.\allowbreak{}2b-\allowbreak{}lora}} & \texttt{7d73ab6535ea} \\
Locator, promoted, 12 layers at a 375-word window & \href{https://huggingface.co/ErtasAI/qmsum-locator-minilm-l12-w375}{\texttt{ErtasAI/\allowbreak{}qmsum-\allowbreak{}locator-\allowbreak{}minilm-\allowbreak{}l12-\allowbreak{}w375}} & \texttt{f54ed53f816b} \\
Locator, protocol-exact, 6 layers at a 900-word window & \href{https://huggingface.co/ErtasAI/qmsum-locator-minilm-l6-w900}{\texttt{ErtasAI/\allowbreak{}qmsum-\allowbreak{}locator-\allowbreak{}minilm-\allowbreak{}l6-\allowbreak{}w900}} & \texttt{a1b9e64d82ca} \\
Span-trained SegEnc of Section 7.5, 406M & \href{https://huggingface.co/ErtasAI/qmsum-summarizer-segenc-406m-spans}{\texttt{ErtasAI/\allowbreak{}qmsum-\allowbreak{}summarizer-\allowbreak{}segenc-\allowbreak{}406m-\allowbreak{}spans}} & \texttt{26abdfc1b037} \\
\bottomrule
\end{tabular}
\end{table}

\texttt{scripts/fetch\_artifacts.py} fetches the three pipeline artifacts at these revisions, verifies recorded SHA256 hashes and installs them at default paths. A second machine reproduced all 281 test predictions byte-identically and scores to full float precision.

We do not release the DialogLED fine-tune because \texttt{MingZhong/\allowbreak{}DialogLED-\allowbreak{}large-\allowbreak{}5120} publishes no model licence; its linked MIT licence covers code. Scores, code and configuration remain available, subject to Sections 8.4 and 8.9.

The summarizer adapter inherits LFM Open License v1.0; span-trained SegEnc and copied Socratic predictions carry BSD 3-Clause; locators, code and our predictions are Apache 2.0.

\section*{Use of AI tools}

Agentic coding tools supported implementation, experiment execution and prose drafting under the author's direction. The author designed the protocol, selected conditions and measurements, made promotion and test-touch decisions, and verified all numbers against released per-query files.

No generative tool is credited as an author. Responsibility for the contents of this paper rests with the author.

\clearpage
\bibliographystyle{tmlr}
\bibliography{refs}

\clearpage
\appendix

\section{The prompt}

Every hosted model receives this protocol-fixed template verbatim:

\begin{quote}
\ttfamily\raggedright\frenchspacing
You are given a meeting transcript and a query. Answer the query with a concise summary based only on the transcript.\\[5pt]
Query: \{query\}\\[5pt]
Transcript:\\
\{transcript\}\\[5pt]
Summary:
\end{quote}

Blank lines are part of the template. \texttt{\{query\}} is released QMSum text; \texttt{\{transcript\}} contains one \texttt{speaker: text} utterance per line. System rows use selected windows in transcript order, capped at 2,000 promoted or 3,000 protocol-exact words. Proprietary hosted rows use the full transcript under a nonbinding 40,000-word cap. Section 7.4 uses located spans or the first 4,500 transcript words.

Our fine-tuned system tokenizes the filled template plus one space as a raw completion, with training loss masked to response tokens and an appended end-of-sequence token. DialogLED uses it as encoder input. Hosted and base-model rows receive it as the sole user message through their chat template, with no system message (Sections 5.1 and 7.4).

Two families of rows use their own input formats, both described where the rows are introduced. The community checkpoints of Section 5.2 take \texttt{query </s> transcript} inputs matching their published fine-tuning convention. The SegEnc rows use that architecture's own query encoding: the Section 5.4 row is scored from the authors' released predictions, so we ran no inference for it, and the span-trained variant of Section 7.5 reads our located spans through the same query-encoding mechanism. The locator consumes query and window text pairs directly, with no prompt template involved (Section 4.1).

The template lives in the protocol module of the released code, and a frozen-protocol test asserts it byte for byte, so a drifted copy fails the test suite before it can produce a row.
\end{document}